\documentclass[10pt,twocolumn,letterpaper]{article}

\usepackage{times}
\usepackage{epsfig}
\usepackage{graphicx}
\usepackage{amsmath}
\usepackage{amssymb}

\usepackage[]{caption,subfig}
\usepackage{mdwlist}
\usepackage[lined, boxed]{algorithm2e}
\usepackage{multirow}

\graphicspath{{figs/}}

\usepackage[pagebackref=true,breaklinks=true,letterpaper=true,colorlinks,bookmarks=false]{hyperref}

\begin{document}

\title{Domain-invariant Face Recognition using Learned Low-rank Transformation}

\author{Qiang Qiu\\
Duke University\\
Durham, NC, 27708\\
{\tt\small qiang.qiu@duke.edu}
\and
Guillermo Sapiro\\
Duke University\\
Durham, NC, 27708\\
{\tt\small guillermo.sapiro@duke.edu}
\and
Ching-Hui Chen\\
University of Maryland\\
College Park, MD 20742\\
{\tt\small  ching@umd.edu}
}

\maketitle

\begin{abstract}
We present a low-rank transformation approach to compensate for face variations due to changes in visual domains, such as pose and illumination. The key idea is to learn discriminative linear transformations for face images using matrix rank as the optimization criteria.
The learned linear transformations restore a shared low-rank structure for faces from the same subject, and, at the same time, force a high-rank structure for faces from different subjects.
In this way, among the transformed faces, we reduce variations caused by domain changes within the classes, and increase separations between the classes for better face recognition across domains.
Extensive experiments using public datasets are presented to demonstrate the effectiveness of our approach for face recognition across domains. The potential of the approach for feature extraction in generic object recognition and coded aperture design are discussed as well.
\end{abstract}

\section{Introduction}

\begin{figure*} [t]
\centering
\includegraphics[angle=0, height=0.43\textwidth, width=.70\textwidth]{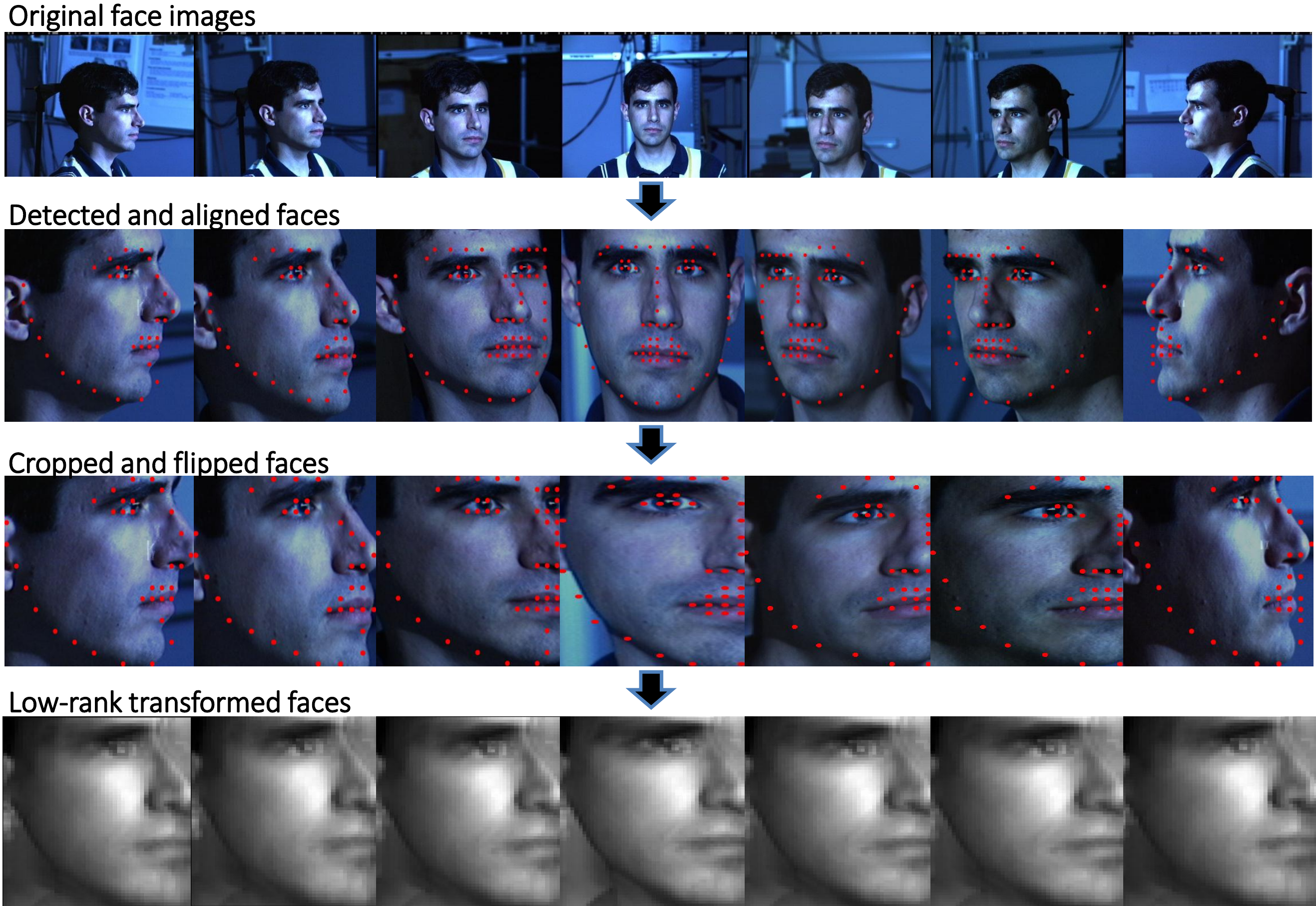}
\caption{Face recognition across pose through learned low-rank transformation.  In the second row, the input faces are first detected and aligned, e.g., using the method in \cite{posemodel}. Pose models defined in \cite{posemodel} enable an optional crop-and-flip step to retain the more informative side of a face in the third row. Our proposed approach learns linear transformations for face images to restore for the same subject a low-dimensional structure as shown in the last row. By comparing the last row to the first row, we can easily notice that faces from the same subject across different poses are more visually similar in the new transformed space, enabling better face recognition across pose (note that the goal is recognition and not reconstruction).}
\label{fig:overview}
\end{figure*}

Face images from the same subject can be well-approximated by a low-dimensional subspace \cite{9point}, \cite{Wright09}.
Under the assumption of Lambertian reflectance, \cite{9point} shows that images of an object obtained under a wide variety of lighting conditions can be approximated accurately with a 9-dimensional linear subspace.
Motivated by \cite{9point}, a sparse representation-based classification (SRC) method is proposed in \cite{Wright09} based on the observation that a face image approximately lies in the linear span of faces from the same subject. SRC has demonstrated the state-of-the-art performance in face recognition.
Thus, we arrange face images from the same subject as columns of a single matrix, and this matrix should be approximately \emph{low-rank}. However, this low-rank structure is often violated for realistic face images.
First, faces are not perfectly Lambertian, and exhibit cast shadows and specularities \cite{rpca}. Second,
real face images are often captured in different visual domains, e.g., under different pose and illumination conditions. As indicated in \cite{src-align}, face recognition methods based on the low-rank assumption, e.g., SRC, often do not deal with misalignment and require a rich set of illuminations in the training.

Recent efforts have been invested in seeking transformations such that the transformed images can be decomposed as the sum of a low-rank matrix component and a sparse error matrix one \cite{RASL, lrsalient, TILT}. \cite{RASL} and \cite{TILT} are proposed for image alignment, and \cite{lrsalient} is discussed in the context of salient object detection. All these methods build on recent theoretical and computational advances in rank minimization.

In this paper, we introduce learned image low-rank transformations for face recognition across domains.  We propose to learn linear discriminative transformations to restore for faces from the same subject the low-dimensional structure that is violated due to domain changes, and, at the same time, to force a high-rank structure for faces from different subjects.
Intuitively, the proposed method shares some of the attributes of the LDA method, but with significantly different metric. Similar to LDA, our method reduces the variation within the classes and increases separations between the classes to achieve better recognition. However, we adopt matrix rank as the key criterion to learn optimal class-based transformations for each subject, or a single global transformation over all subjects.

Real world face recognition applications often require an automatic process of face detection, alignment, and recognition. Various popular face detection methods, e.g., Viola-Jones detector \cite{Viola-Jones}; and face alignment methods, e.g., Congealing \cite{congeal}; have been shown practical for realistic face images. However, many of these methods often experience severe performance degradation given large pose (domain) variations among face images.
As one of the very recent efforts, \cite{posemodel} presents a promising framework in dealing with face detection and alignment across poses, adopting a deformable part model based approach \cite{DPM} to learn various pose models for faces. As shown in the second row of Fig.~\ref{fig:overview}, a given face image is detected and aligned to the closet pose model. State-of-the-art face detection and alignment performance is reported in \cite{posemodel} for real faces with significant pose variations. However, how to recognize the detected and aligned faces across poses is still left to be addressed.

Face recognition across domains, e.g., pose and illumination, has proved to be a challenging problem \cite{s-smd, lightfield,dadl,tensorface1}. In \cite{lightfield}, the eigen light-field (ELF) algorithm is presented for face recognition across pose and illumination. This algorithm operates by estimating the eigen light field or the plenoptic function of the subject's head using all the pixels of various images. In \cite{s-smd, smd}, face recognition across pose is performed using stereo matching distance (SMD). The cost to match a probe image to a gallery image is used to evaluate the similarity of the two images. In \cite{dadl},
a domain adaptive dictionary learning (DADL) framework is proposed to transform a dictionary learned from one visual domain to the other, while maintaining a domain-invariant sparse representation of a signal. This enhanced sparse representation based approach outperforms the SRC method \cite{Wright09} for face recognition across domains.

We propose to address face recognition across domains through learned image low-rank transformation.
As shown in Fig.~\ref{fig:overview}, our approach expects the detected and aligned faces, e.g., using \cite{posemodel}.
Pose models defined in \cite{posemodel} enable a simple crop-and-flip step to retain the more informative side of a face.
Our approach learns linear transformations for face images to restore for the same subject a low-dimensional structure.
By comparing the last row to the first row in Fig.~\ref{fig:overview}, we can easily notice that faces from the same subject  across different poses are more visually similar in the new transformed space, enabling better face recognition across pose.

This paper makes the following contributions:
\begin{itemize*}
  \item Image low-rank transformations are discussed in the context of image classification;
  \item A discriminative low-rank transformation approach is proposed to reduce the variation within the classes and increase separations between the classes;
  \item We demonstrate through extensive experiments the significant performance improvements in face recognition across pose and illumination by using our proposed approach.
  \end{itemize*}

 In Section~\ref{sec:lrt}, we formulate the low-rank transformation learning problem for face images.
We discuss both class-based transformation per subject, and global transformation over all subjects to compensate for face variations due to domain shifts.
Experimental evaluations are given in Section~\ref{sec:expr} on public datasets.
Finally, Section~\ref{sec:con} concludes the paper.

\section{Domain-invariant Face Recognition}
\label{sec:lrt}

Given a set of face images from $N$ different subjects (classes), each face image is represented in a $d$-dim feature space, $\mathbf{Y_i}=[\mathbf{y_{i1}}, \ldots, \mathbf{y_{iK}}]$, $\mathbf{y_{ik}} \in \mathbb{R}^d$, denotes the set of $K$ faces of the $i^{th}$ subject. We denote all face images from $N$ different classes as $\mathbf{Y}= [\mathbf{Y_1},  \ldots , \mathbf{Y_N}]$, and all faces except the $i^{th}$ subject as
$\mathbf{Y_{\neg i}} = \mathbf{Y}  \setminus \mathbf{Y_i}$.

Face images from the same subject can be well-approximated by a low-dimensional subspace \cite{9point}, \cite{Wright09}.
Thus, the matrix $\mathbf{Y_i}$ is expected to be \emph{low-rank}, and such low-rank structure is critical to accurate face recognition. However, this low-rank structure is often violated for realistic face images, due to pose and illumination variations.

Our proposed approach learns linear transformations of face images. Such linear transformations restore a low-rank structure for faces from the same subject, and, at the same time, encourage a high-rank structure for faces from different subjects. In this way, the proposed linear transformations help to eliminate the variation within the classes and introduce separations between the classes. Both aspects are critical to achieve accurate face recognition across domains.

\subsection{Low-rank Transformation (LRT) for Faces}

We adopt matrix rank as the key criterion to learn optimal class-based transformations for each
subject, or a single global transformation over all subjects.

\subsubsection{Class-based Transformation per Subject}

The problem to compute individual class-based linear transformation per subject can be formulated as (\ref{itran_obj}).

\begin{align} \label{itran_obj}
\underset{\{\mathbf{T_i}\}_{i=1}^N} \arg \min \sum_{i=1}^N [ ||\mathbf{T_i Y_i}||_* - \lambda||\mathbf{T_i Y_{\neg i}}||_* ],
\end{align}
where $\mathbf{T_i} \in \mathbb{R}^{d \times d}$ denotes the face transformation for the $i^{th}$ subject, and $||\mathbf{\cdot}||_*$ denotes the nuclear norm.  Intuitively, minimizing the first \emph{representation} term $||\mathbf{T_i Y_i}||_*$ results in a consistent representation in the transformed space for the $i^{th}$ subject faces; and minimizing the second \emph{discrimination} term $-||\mathbf{T_i Y_{\neg i}}||_*$ encourages a diverse representation for transformed faces from other subjects for better discrimination.
The parameter $\lambda \ge 0$  regularizes the emphasis on representation or discrimination.

\subsubsection{Global Transformation over All Subjects}

The problem to compute one single global linear transformation over all subjects can be formulated as (\ref{stran_obj}).

\begin{align} \label{stran_obj}
\underset{\mathbf{T}} \arg \min \frac{1}{N} \sum_{i=1}^N ||\mathbf{T Y_i}||_* - \lambda||\mathbf{T Y}||_*,
\end{align}
where $\mathbf{T} \in \mathbb{R}^{d \times d}$ denotes one global face transformation for all classes.
Minimizing the first term $\frac{1}{N} \sum_{i=1}^N ||\mathbf{T Y_i}||_*$ encourages a consistent representation for the transformed faces from the same class; and minimizing the second term $-||\mathbf{T Y}||_*$ encourages a diverse representation for transformed faces from different classes. The parameter $\lambda \ge 0$  weights the discrimination term.

Before proceeding it is interesting to note the connection of this work with recent results on the design of optimal coded apertures for compressed sensing, see \cite{CS1, CS2} and references therein. With a completely different goal in mind, the main criteria there is mutual information, and the learned matrix (mask) has less rows than columns. It is interesting then to note that we could consider the learned matrix in (1) or (2) also as a compression mechanism, by reducing the member of rows, and as such investigate the use of nuclear norm as here exploited for the design of optimal coded apertures for classification. This connection will be exploited in the future.

\subsection{Gradient Descent Low-rank Learning}

Given any matrix $\mathbf{A}$ of rank at most $r$,  the matrix norm $||\mathbf{A}||$ is equal to its largest singular value, and the nuclear norm $||\mathbf{A}||_*$ is equal to the sum of its singular values. Thus, these two norms are related by the following inequality,

\begin{align} \label{norm_ineq}
||\mathbf{A}|| \le ||\mathbf{A}||_* \le r||\mathbf{A}||,
\end{align}

We use gradient descent (though other modern nuclear norm optimization techniques could be considered) to search for the $i^{th}$ class (subject) optimal transformation matrix $\mathbf{T_i}$ that minimizes (\ref{itran_obj}). The partial derivative of (\ref{itran_obj}) w.r.t $\mathbf{T_i}$ is written as,

\begin{align} \label{itran_der}
\frac{\partial}{\partial \mathbf{T_i}}[||\mathbf{T_i Y_i}||_* - \lambda||\mathbf{T_i Y_{\neg i}}||_* ],
\end{align}

Due to the property (\ref{norm_ineq}), by minimizing the matrix norm, one can also minimize an upper bound to the nuclear norm. (\ref{itran_der}) can now be evaluated as,

\begin{align} \label{itran_sub}
\Delta \mathbf{T_i} = \partial ||\mathbf{T_i Y_i}||\mathbf{Y_i}^T - \lambda \partial ||\mathbf{T_i Y_{\neg i}}||\mathbf{Y_{\neg i}}^T,
\end{align}
where $\partial ||\mathbf{\cdot}||$ is the subdifferential of a norm $||\mathbf{\cdot}||$. Given a matrix $\mathbf{A}$, the subdifferential $\partial ||\mathbf{A}||$ can be evaluated using a simple approach shown in Algorithm~\ref{subdifferential} \cite{subdifferential}. By evaluating $\Delta \mathbf{T_i}$, the optimal transformation matrix $\mathbf{T_i}$ can be searched with gradient descent $\mathbf{T_i}^{(t+1)} = \mathbf{T_i}^{(t)} + \nu \Delta \mathbf{T_i}$ ($\nu > 0$ defines the step size.). After each iteration, we normalize $\mathbf{T_i}$ as $\frac{\mathbf{T_i}}{||\mathbf{T_i}||}$. This algorithm guarantees convergence to a local minimum. (\ref{stran_obj}) can be optimized in a similar way.

\begin{algorithm}[ht]
\footnotesize
\KwIn{an $m \times n$ matrix $\mathbf{A}$, a small threshold value $\delta$}
\KwOut{the subdifferential of the matrix norm $\partial ||\mathbf{A}||$.}
\Begin{
\BlankLine
1. Perform singular value decomposition: \\
$\mathbf{A}=\mathbf{U} \mathbf{\Sigma} \mathbf{V}$ \;
\BlankLine
2. $s \leftarrow$ the number of singular values smaller than $\delta$ , \\
3. Partition $\mathbf{U}$ and $\mathbf{V}$ as \\
$\mathbf{U} = [\mathbf{U}^{(1)}, \mathbf{U}^{(2)}]$, $\mathbf{V} = [\mathbf{V}^{(1)}, \mathbf{V}^{(2)}]$ \;
where $\mathbf{U}^{(1)}$ and $\mathbf{V}^{(1)}$ have $(n-s)$ columns. \\
\BlankLine
4. Generate a random matrix $\mathbf{B}$ of the size $(m-n+s)\times s$, \\
$\mathbf{B} \leftarrow \frac{\mathbf{B}}{||\mathbf{B}||}$ \;
\BlankLine
5. $\partial ||\mathbf{A}|| \leftarrow \mathbf{U}^{(1)} \mathbf{V}^{(1)T} + \mathbf{U}^{(2)} \mathbf{B} \mathbf{V}^{(2)T}$ \;
\BlankLine
6. return $\partial ||\mathbf{A}||$ \;

}
\caption{An approach to evaluate the subdifferential of a matrix norm.}
\label{subdifferential}
\end{algorithm}

\subsection{Face Recognition using LRT}

After a global transformation matrix $\textbf{T}$ is learned, we can perform face recognition in the transformed space by simply considering the transformed faces $\mathbf{TY}$ as the new features. For example, when a Nearest Neighbor (NN) classifier is used, a testing face $\mathbf{y}$ uses $\mathbf{Ty}$ as the feature and searches for nearest neighbors among $\mathbf{TY}$.

To fully exploit the low-rank structure of the transformed faces, we propose to perform recognition through the following procedure:

\begin{itemize*}
\item For the $i^{th}$ class, we first recover its low-rank representation $\mathbf{L_i}$ by performing low-rank decomposition (\ref{rpca}), e.g., using RPCA \cite{rpca}.
    \begin{align} \label{rpca}
\underset{\mathbf{L_i}, \mathbf{S_i}} \arg \min ||\mathbf{L_i}||_* + \beta ||\mathbf{S_i}||_1 ~~s.t.~ \mathbf{TY_i} =\mathbf{L_i}+\mathbf{S_i}.
\end{align}
\item Each testing image $\mathbf{y}$ will then be assigned to the low-rank subspace $\mathbf{L_i}$ that gives the minimal reconstruction error through sparse decomposition (\ref{omp}), e.g., using OMP \cite{omp},
\begin{align} \label{omp}
\underset{\mathbf{x}} \arg \min \|\mathbf{Ty}-\mathbf{L_ix}\|_{2}^{2} ~~s.t.~ \|\mathbf{x}\|_{0}\leq T ,
\end{align}
where $T$ is a predefined sparsity value.
\end{itemize*}

When class-based transformations $\{\mathbf{T_i}\}_{i=1}^N$ are learned, we perform recognition in a similar way as discussed above. However, now we need to iteratively apply all $\mathbf{T_i}$ to each testing image and then pick the best one.

\section{Experimental Evaluation}
\label{sec:expr}

This section presents experimental evaluations on two public face datasets: the CMU PIE dataset \cite{pie} and the Extended YaleB dataset \cite{yaleb}. The PIE dataset consists of 68 subjects imaged simultaneously under 13 different poses and 21 lighting conditions.
The Extended YaleB dataset contains 38 subjects with near frontal pose under 64 lighting conditions.
All the face images are resized to $20 \times 20$. We adopt a NN classifier unless otherwise specified.

\subsection{Illustrative Examples of LRT}
\label{sec:usps}

\begin{figure*} [t]
\centering
\subfloat[Class-based transformation with $\lambda = 0$ (recognition rate = $96.95\%$).] {\label{fig:uspsClassB0} \includegraphics[angle=0, height=0.15\textwidth, width=.9\textwidth]{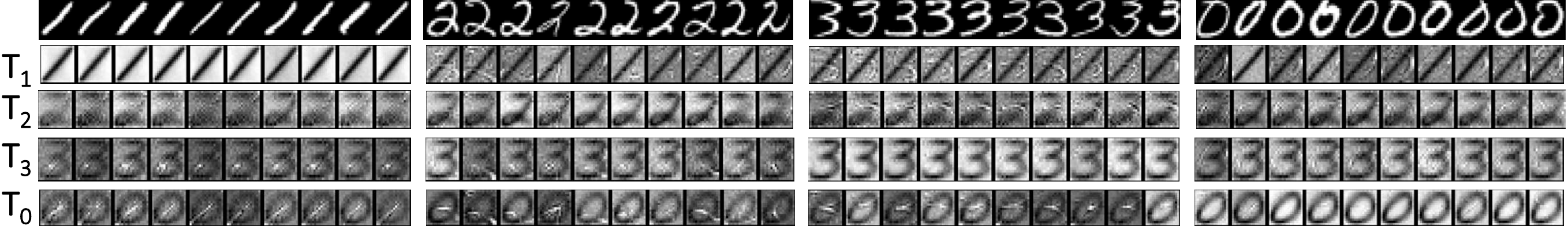} \hspace{0pt}} \\
\subfloat[Class-based transformation with $\lambda = 0.8$ (recognition rate = $98.68\%$).] {\label{fig:uspsClassB1} \includegraphics[angle=0, height=0.15\textwidth, width=.9\textwidth]{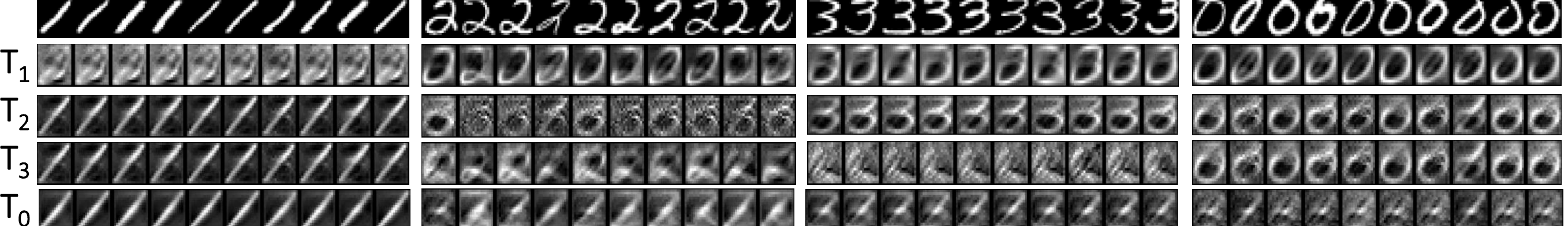}} \\
\subfloat[Global transformation with $\lambda = 0$ (recognition rate = $96.27\%$).] {\label{fig:uspsShareB0} \includegraphics[angle=0, height=0.06\textwidth, width=.87\textwidth]{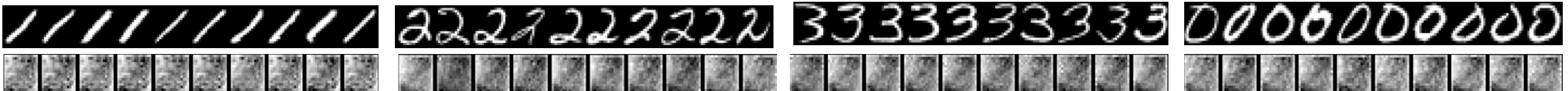} \hspace{0pt}} \\
\subfloat[Global transformation with $\lambda = 0.8$ (recognition rate = $97.68\%$).] {\label{fig:uspsShareB1} \includegraphics[angle=0, height=0.06\textwidth, width=.87\textwidth]{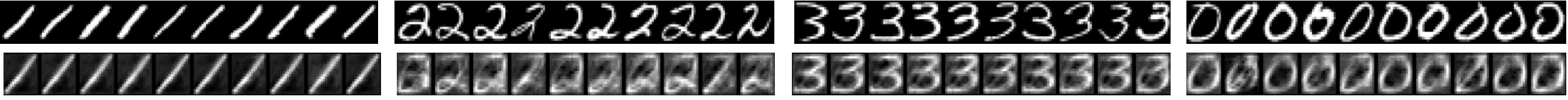}}
\caption{Illustrative examples of low-rank transformed images.}
\label{fig:usps}
\end{figure*}

We first choose a non-face dataset to enable better visualized illustrations of the proposed low-rank transformation.
This also helps to illustrate how the proposed framework has the potential to be integrated into generic object recognition techniques.
We conduct a set of experiments on the first four digits, i.e., $\{0,1,2,3\}$, in the USPS digit dataset.
Half of the data are used for training and the rest is used for testing.
This example is only for pedagogic reasons and not intended to present state of the art digit recognition.

We learn image transformations from the training data based on (\ref{itran_obj}) and (\ref{stran_obj}) with various $\lambda$ values. We then randomly select from the testing data 10 samples per class, and visualize the transformed images in Fig.~\ref{fig:usps}.

Fig.~\ref{fig:uspsClassB0} and Fig.~\ref{fig:uspsClassB1} illustrate the class-based transformation in (\ref{itran_obj}). We learn from the training data one transformation matrix per digit class, and denote them as $\mathbf{T_0}$, $\mathbf{T_1}$, $\mathbf{T_2}$ and $\mathbf{T_3}$ respectively.
The first row shows the randomly selected test samples, and the subsequent rows show the
transformed images using the respective $\mathbf{T_i}$.

In Fig.~\ref{fig:uspsClassB0}, we set $\lambda$ to $0$ to use only the first low-rank representation term in (\ref{itran_obj}).  As we can notice from the second row of the first column, the third row of the second column, and so on,  $\mathbf{T_i}$ enforces a consistent intra-class low-rank representation for the respective $i^{th}$ class.

In Fig.~\ref{fig:uspsClassB1}, we set $\lambda$ to a relatively large value to emphasize the second discrimination term in (\ref{itran_obj}). Now we observe that each class exhibits more diverse representation from the other classes, e.g., by comparing the first column to the rest in the second row, the second column to the rest in the third row, and so on. We  also observe that $\mathbf{T_i}$ still enforces a consistent intra-class description for the respective class with significantly weaker reconstruction power, as expected.

Fig.~\ref{fig:uspsShareB0} and Fig.~\ref{fig:uspsShareB1} illustrate the global transformation in (\ref{stran_obj}).
We learn from the training data one global transformation matrix for all four digit classes. We have similar observations as discussed above: the first term in (\ref{stran_obj}) enforces a consistent low-rank intra-class representation, and
the second term in (\ref{stran_obj}) encourages a diverse inter-class representation. However, we notice that, for the global transformation, it is the second term that enables better reconstruction.

In both scenarios, the inclusion of the discriminative term ($\lambda >0$) improves performance.

\subsection{Face Recognition across Illumination}

\begin{figure} [ht]
\centering
 \includegraphics[angle=0, height=0.25\textwidth, width=.35\textwidth]{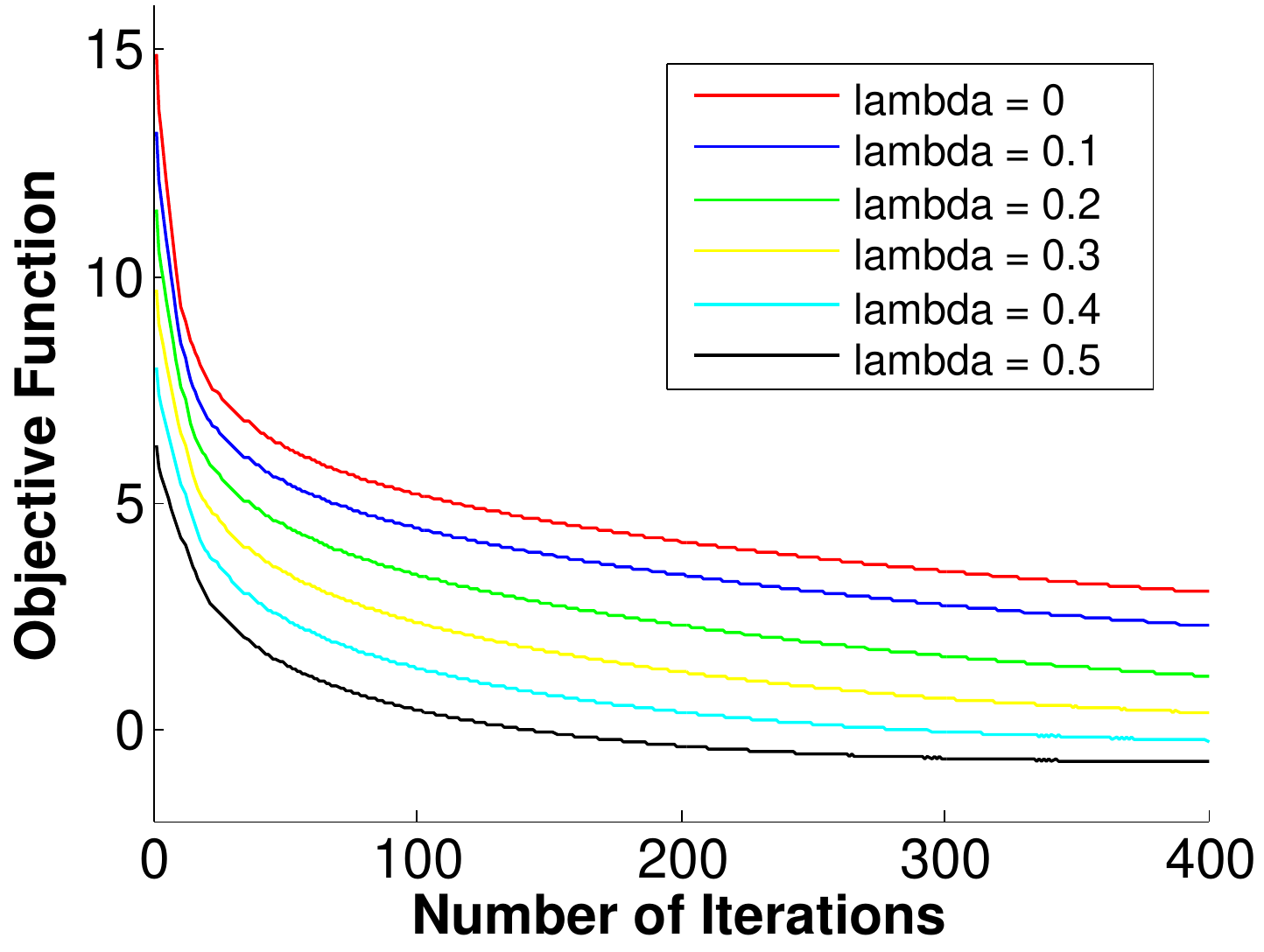}
\caption{ Convergence of the objective function with various $\lambda$ values.}
\label{fig:Yaleconverge}
\end{figure}

\begin{table}[h]
\centering
{\small
	\begin{tabular}{|l|l|}
	\hline
Method & Accuracy (\%) \\
	\hline
 \hline
D-KSVD \cite{Zhang10} & 94.10 \\
LC-KSVD \cite{lcksvd} & 96.70 \\
SRC \cite{Wright09} & 97.20 \\
\hline
\hline
Original+NN & 91.77 \\
Class LRT+NN & 97.86 \\
Class LRT+OMP  & 92.43 \\
Global LRT+NN & 99.10 \\
Global LRT+OMP  & \textbf{99.51} \\
\hline
	\end{tabular}
}	
	\caption{Recognition accuracies (\%) under illumination variations for the Extended YaleB dataset. The recognition accuracy is increased from $91.77\%$ to $99.10\%$ by simply applying the learned transformation matrix to the original face images.}
	\label{tab:yaleacc}
\end{table}

\begin{figure*} [t]
\centering
\subfloat[Low-rank decomposition of globally transformed training samples] {\label{fig:YaleShareTrain} \includegraphics[angle=0, height=0.29\textwidth, width=.9\textwidth]{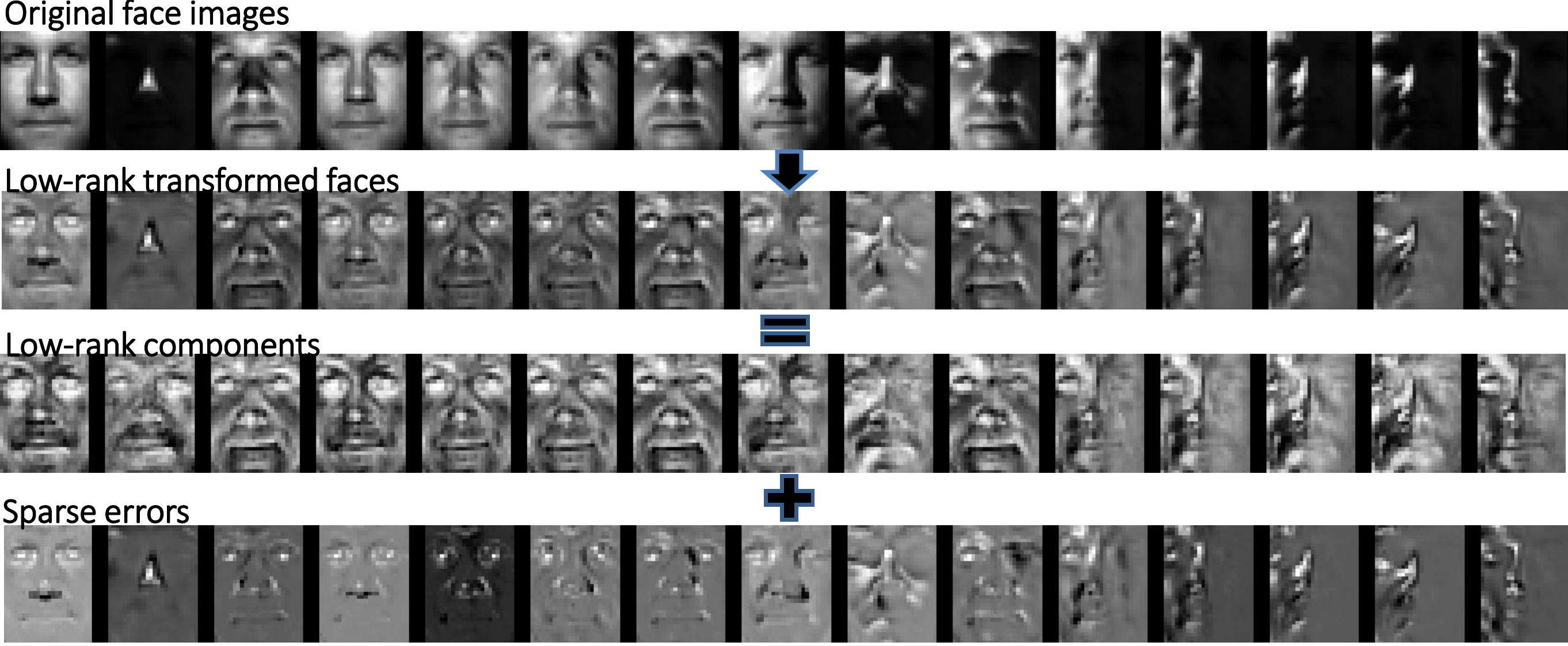} \hspace{0pt}} \\
\subfloat[Globally transformed testing samples] {\label{fig:YaleShareTest} \includegraphics[angle=0, height=0.15\textwidth, width=.9\textwidth]{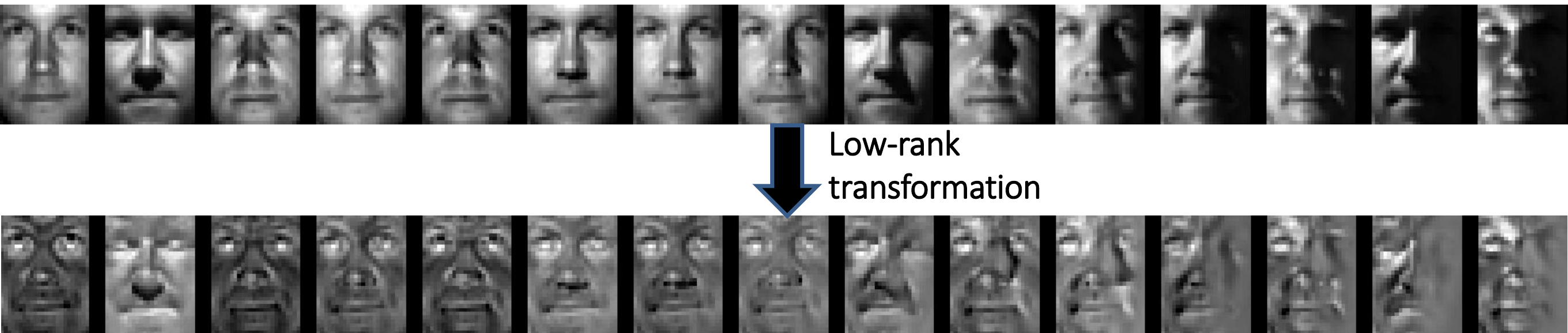}} \\
\subfloat[Mean low-rank components for subjects in the training data] {\label{fig:YaleSTavglowrank} \includegraphics[angle=0, height=0.08\textwidth, width=.9\textwidth]{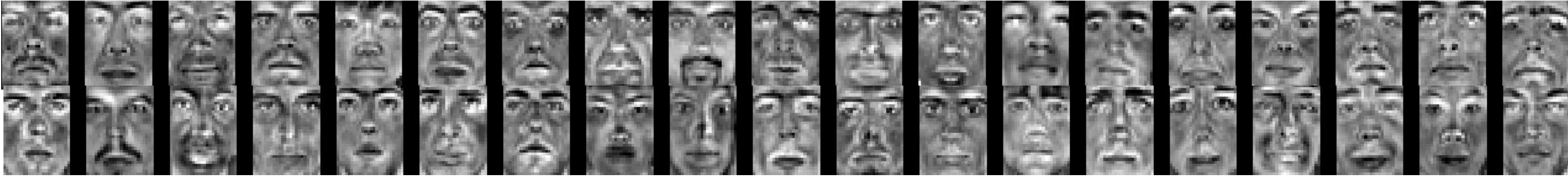}}
\caption{Face recognition across illumination using global low-rank transformation.}
\label{fig:vt}
\end{figure*}

For the Extended YaleB dataset, we adopt a similar setup as described in \cite{lcksvd,Zhang10}.  We split the dataset into two halves by randomly selecting 32 lighting conditions for training, and the other half for testing.

We learn a global low-rank transformation matrix from the training data. As shown in Fig.~\ref{fig:Yaleconverge}, we always observe empirical convergence for the proposed transformation learning algorithms. In all of our experiments, we adopt $100$ iterations for the gradient descent updates.
As discussed in Section~\ref{sec:usps}, the value of $\lambda$ regularizes the emphasis on the representation and discrimination terms in the objective function (\ref{stran_obj}). We observed the best recognition accuracy is achieved at a balance between these two terms.  In our experiments, we choose $\lambda=0.1$.
In general, the value of $\lambda$ can be estimated through cross-validations.

We report recognition accuracies in Table~\ref{tab:yaleacc}. We make the following observations. First,
 the recognition accuracy is increased from $91.77\%$ to $99.10\%$ by simply applying the learned transformation matrix to the original face images.
 Second, the best accuracy is obtained by first recovering the low-rank subspace for each subject, e.g., the third row in Fig.~\ref{fig:YaleShareTrain}. Then, each transformed testing face,  e.g., the second row in Fig.~\ref{fig:YaleShareTest}, is sparsely decomposed over the low-rank subspace of each subject through OMP, and classified to the subject with the minimal reconstruction error. A sparsity value 10 is used here for OMP.
 As shown in Fig.~\ref{fig:YaleSTavglowrank}, the low-rank representation for each subject shows reduced variations caused by illumination.
 Third, the global transformation performs better here than class-based transformations, which can be due to the fact that illumination in this dataset varies in a globally coordinated way across subjects.
   Last but not least, our method outperforms state-of-the-art sparse representation based face recognition methods.

\subsection{Face Recognition across Pose}

\begin{table}[h]
\centering
{\small
	\begin{tabular}{|l|l|l|l|}
	\hline
Method & Frontal & Side & Profile \\
 & (c27) & (c05) & (c22) \\
	\hline
 \hline
SMD \cite{smd} & 83 & 82 & 57 \\
\hline
\hline
Original+NN & 39.85 & 37.65 & 17.06 \\
Original(crop+flip)+NN & 44.12 & 45.88 & 22.94 \\
Class LRT+NN &  98.97 & 96.91& 67.65\\
Class LRT+OMP  & \textbf{100} & \textbf{100} & \textbf{67.65}\\
Global LRT+NN & 97.06 & 95.58& 50\\
Global LRT+OMP  & 100 & 98.53& 57.35\\
\hline
	\end{tabular}
}	
	\caption{Recognition accuracies (\%) under pose variations for the CMU PIE dataset.}
	\label{tab:pieacc}
\end{table}

\begin{figure*} [t]
\centering
\subfloat[Low-rank decomposition of class-based transformed training samples for \emph{subject3} ] {\label{fig:PIEClassTrain01} \includegraphics[angle=0, height=0.23\textwidth, width=.5\textwidth]{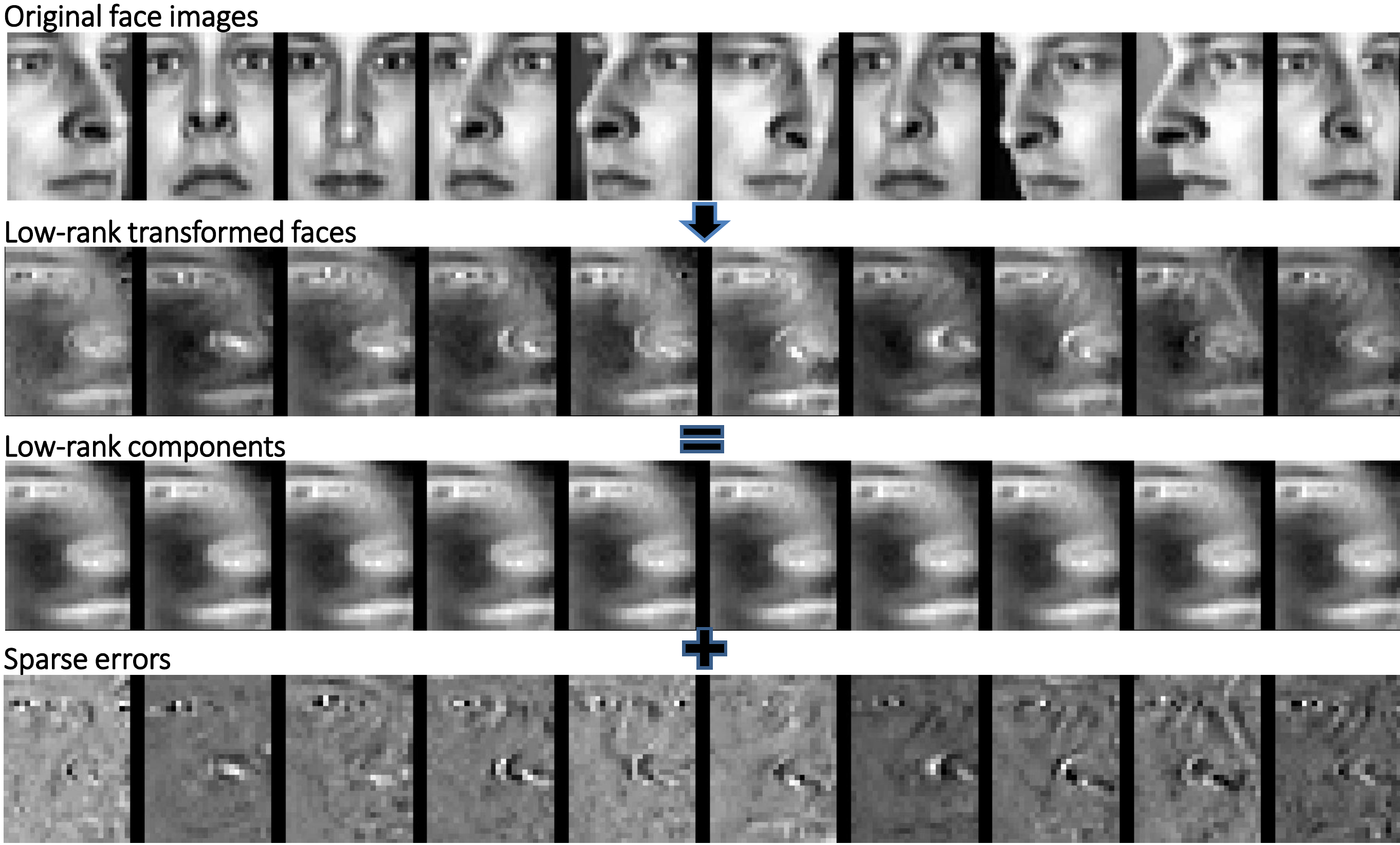} \hspace{0pt}}
\subfloat[Low-rank decomposition of class-based transformed training samples for \emph{subject1} ] {\label{fig:PIEClassTrain02} \includegraphics[angle=0, height=0.23\textwidth, width=.5\textwidth]{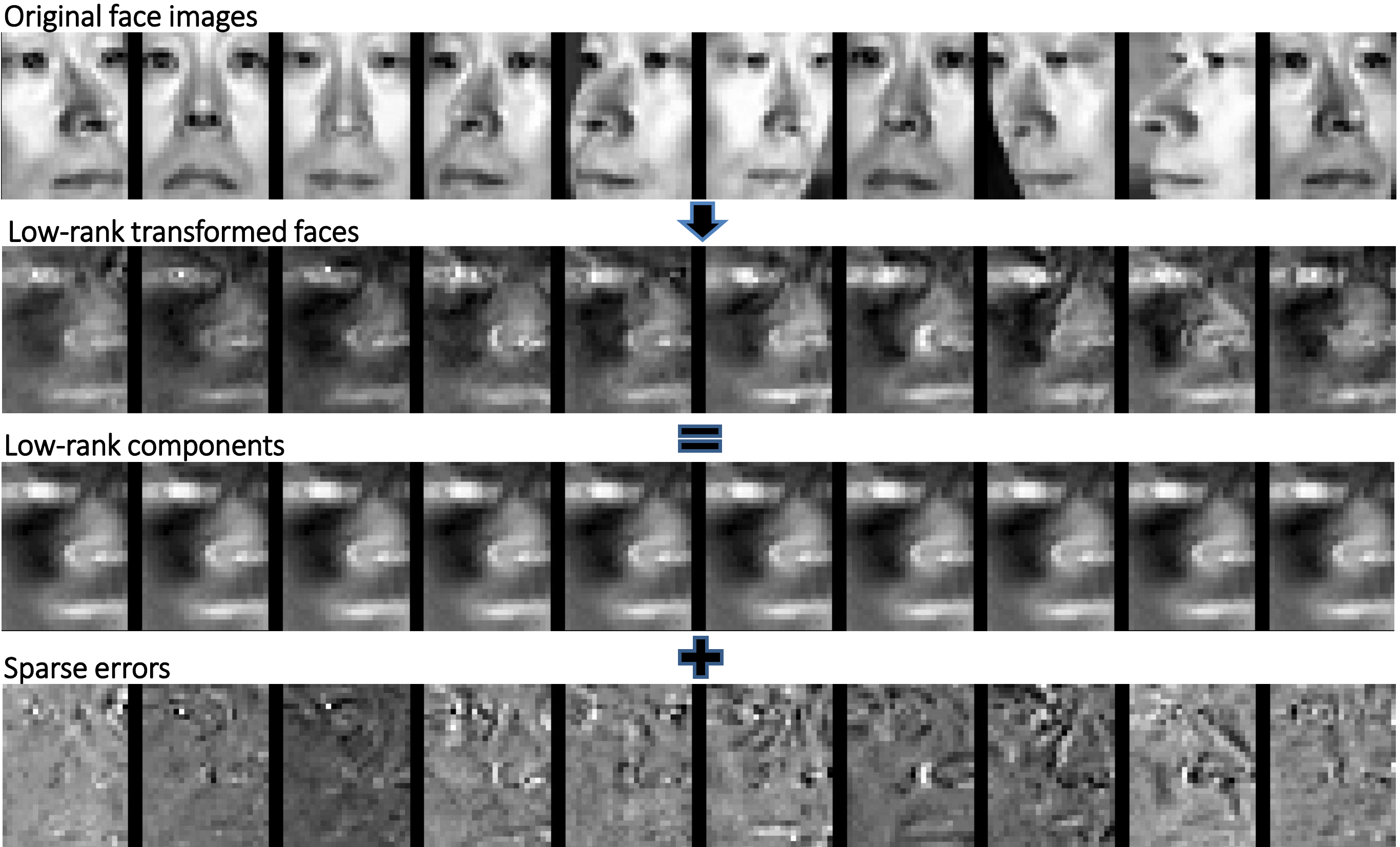} \hspace{0pt}} \\
\subfloat[class-based transformed testing samples for \emph{subject3}] {\label{fig:PIEClassTest01} \includegraphics[angle=0, height=0.06\textwidth, width=.4\textwidth]{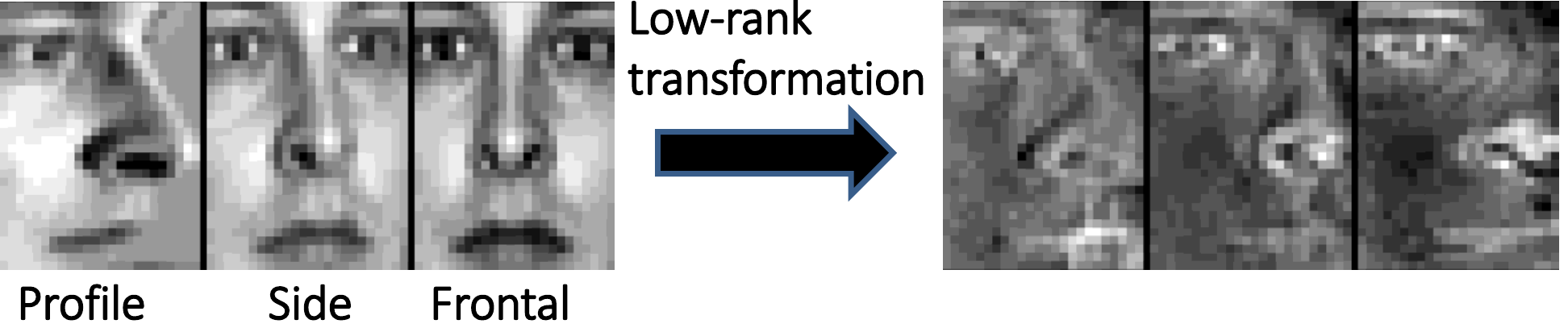} \hspace{10pt}}
\subfloat[class-based transformed testing samples for \emph{subject1}] {\label{fig:PIEClassTest02} \includegraphics[angle=0, height=0.06\textwidth, width=.4\textwidth]{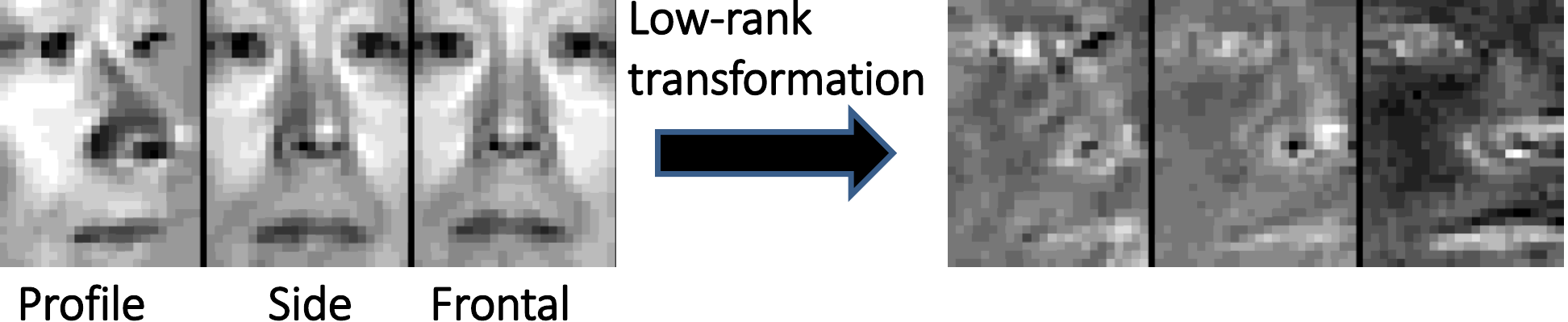}}
\caption{Face recognition across pose using class-based low-rank transformation.}
\label{fig:vt}
\end{figure*}

We adopt the similar setup as described in \cite{smd} to enable the comparison. In this experiment, we classify 68 subjects in three poses, frontal (c27), side (c05), and profile (c22), under lighting condition 12. We use the remaining poses as the training data.

We learn class-based low-rank transformation matrix per subject from the training data.
Table~\ref{tab:pieacc} shows the face recognition accuracies under pose variations for the CMU PIE dataset. We make the following observations.
First, the recognition accuracy is dramatically increased after applying the learned transformations. Here we apply the crop-and-flip step discussed in Fig.~\ref{fig:overview}.
Second, the best accuracy is obtained by recovering the low-rank subspace for each subject, e.g., the third row in Fig.~\ref{fig:PIEClassTrain01} and Fig.~\ref{fig:PIEClassTrain02}. Then, each transformed testing face,  e.g.,  Fig.~\ref{fig:PIEClassTest01} and Fig.~\ref{fig:PIEClassTest02}, is sparsely decomposed over the low-rank subspace of each subject through OMP, and classified to the subject with the minimal reconstruction error.
It is noted that now we need to iteratively apply all $\mathbf{T_i}$ to each testing image and then pick the best one.
Third, the class-based transformation performs better than the global transformation in this case. Therefore, the choice between these two settings is data dependent.
Last but not least, our method outperforms SMD. To the best of our knowledge, SMD reports the best recognition performance in such experimental setup. However, it is noted that SMD is an unsupervised method, and the proposed method requires training.

\subsection{Face Recognition across Illumination and Pose}

\begin{figure*} [t]
\centering
 \subfloat[Pose c02] {\label{fig:xxx} \includegraphics[angle=0, height=0.23\textwidth, width=.25\textwidth]{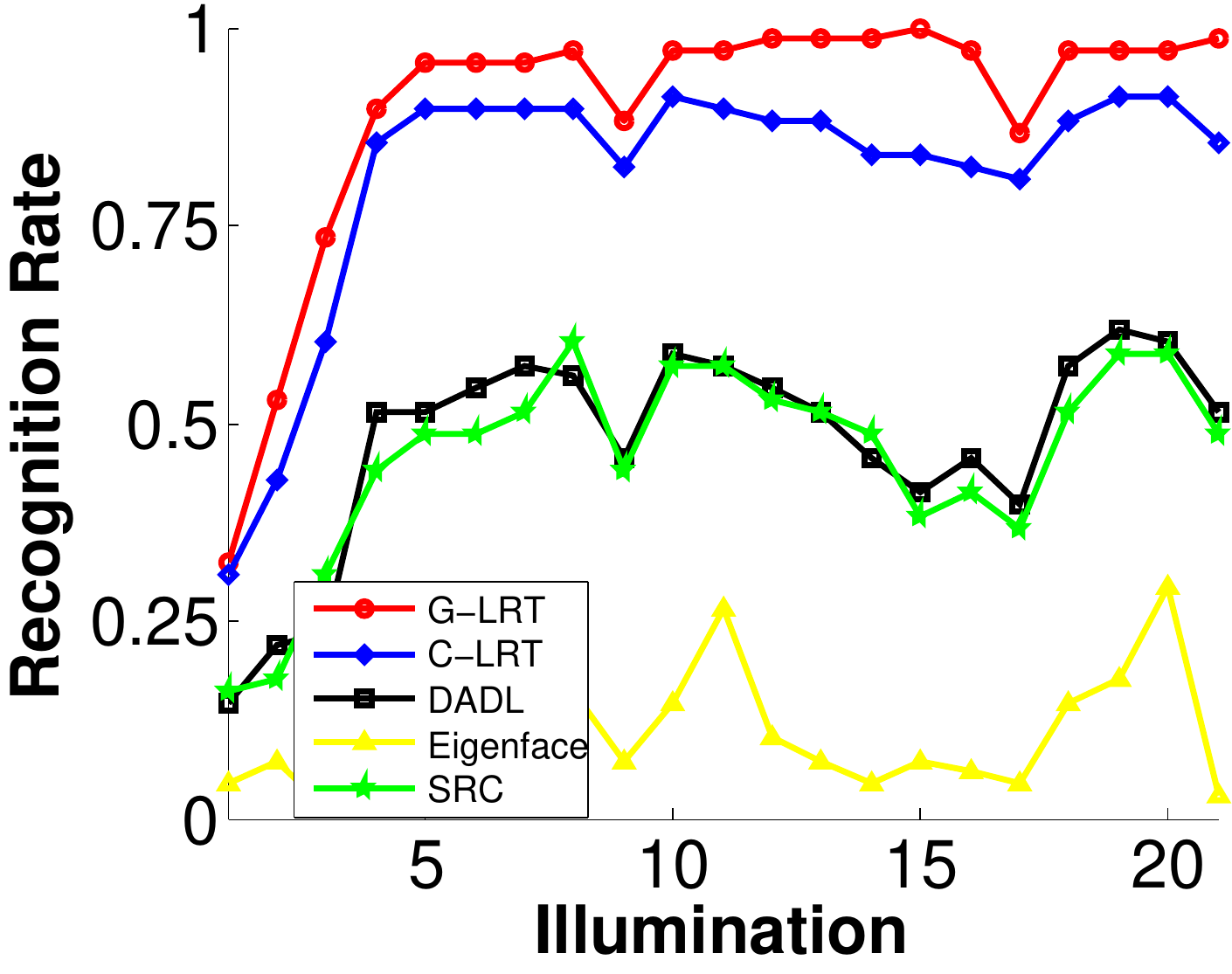} \hspace{0pt}}
  \subfloat[Pose c05] {\label{fig:xxx} \includegraphics[angle=0, height=0.23\textwidth, width=.25\textwidth]{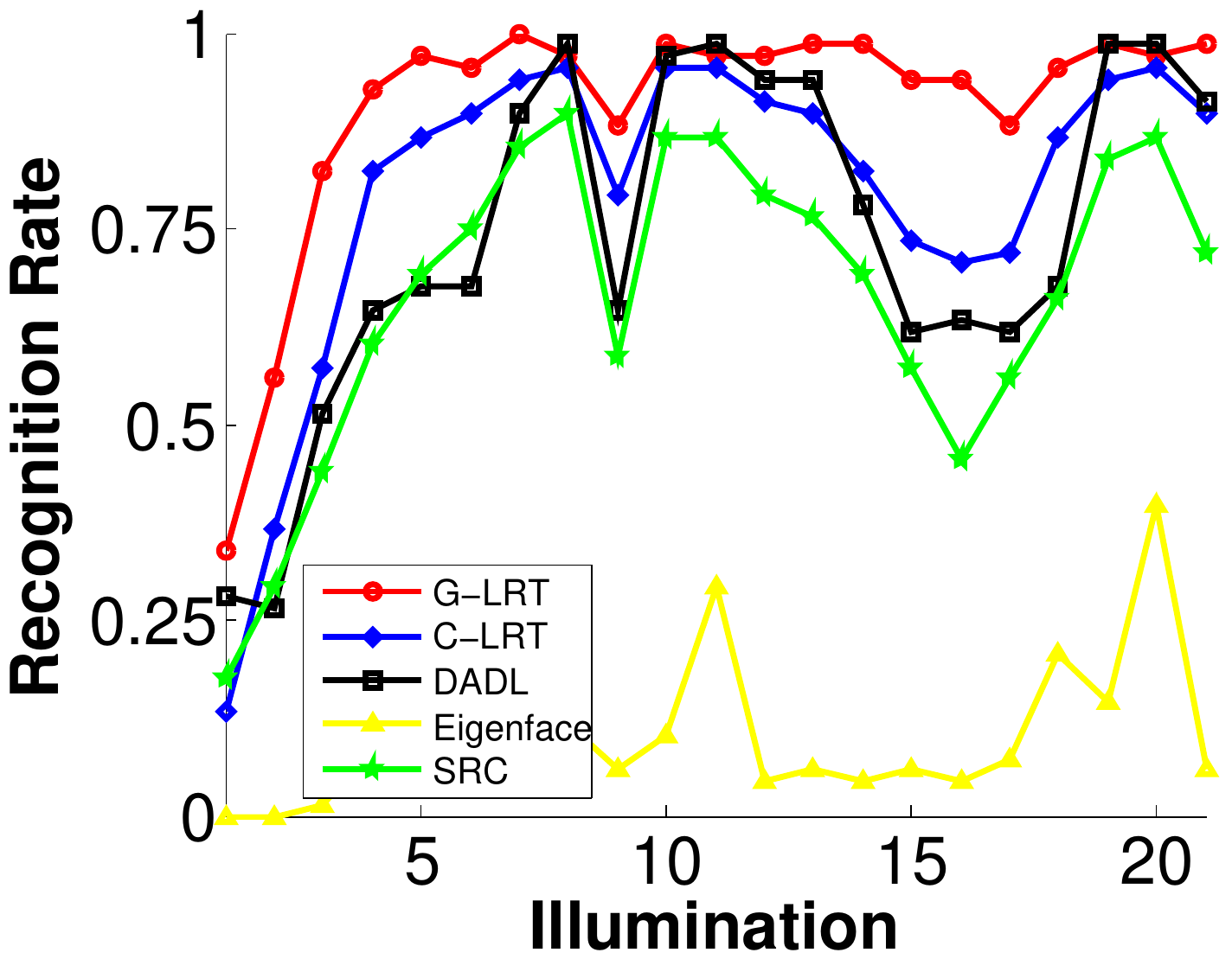}}
    \subfloat[Pose c29] {\label{fig:xxx} \includegraphics[angle=0, height=0.23\textwidth, width=.25\textwidth]{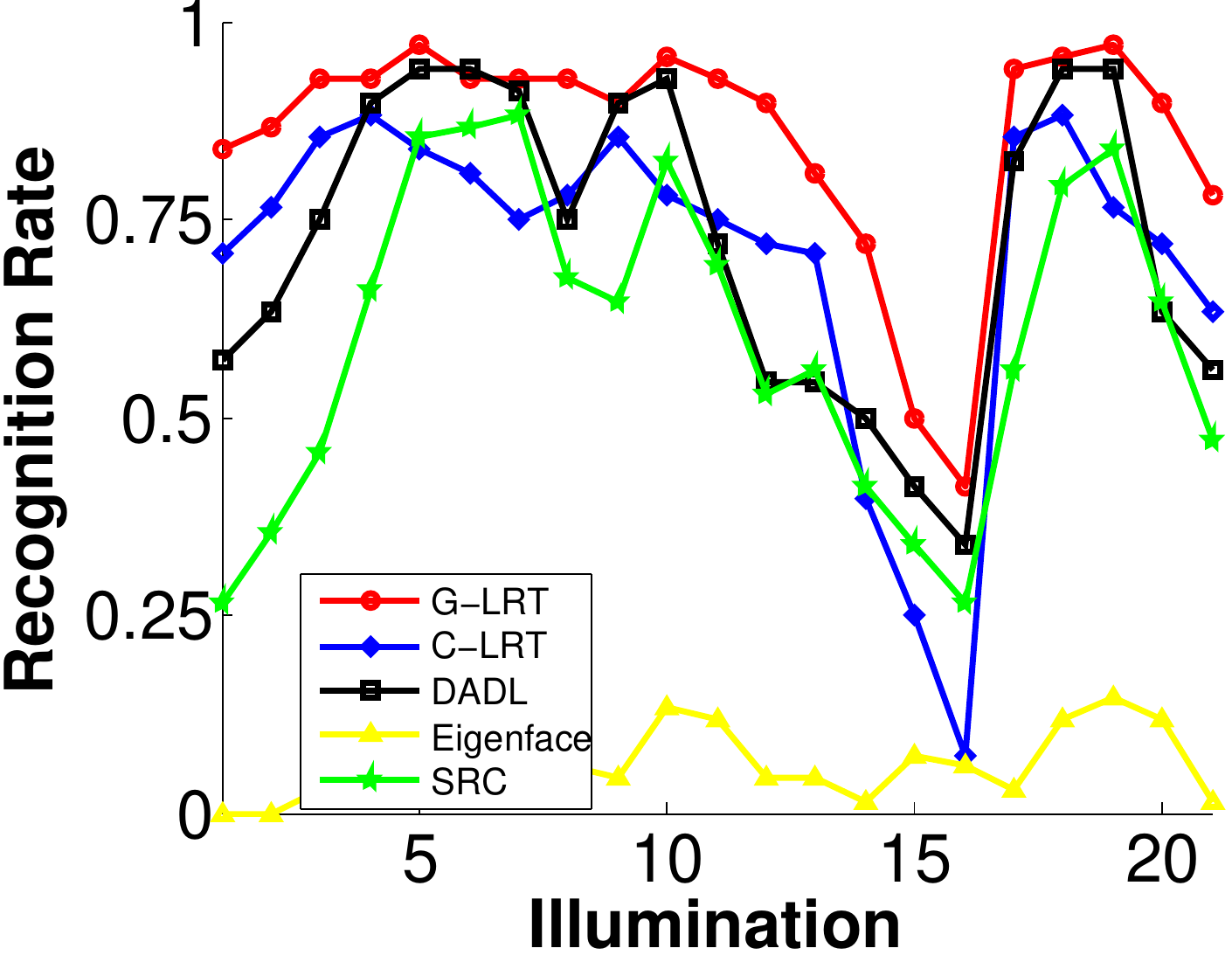}}
  \subfloat[Pose c14] {\label{fig:xxx} \includegraphics[angle=0, height=0.23\textwidth, width=.25\textwidth]{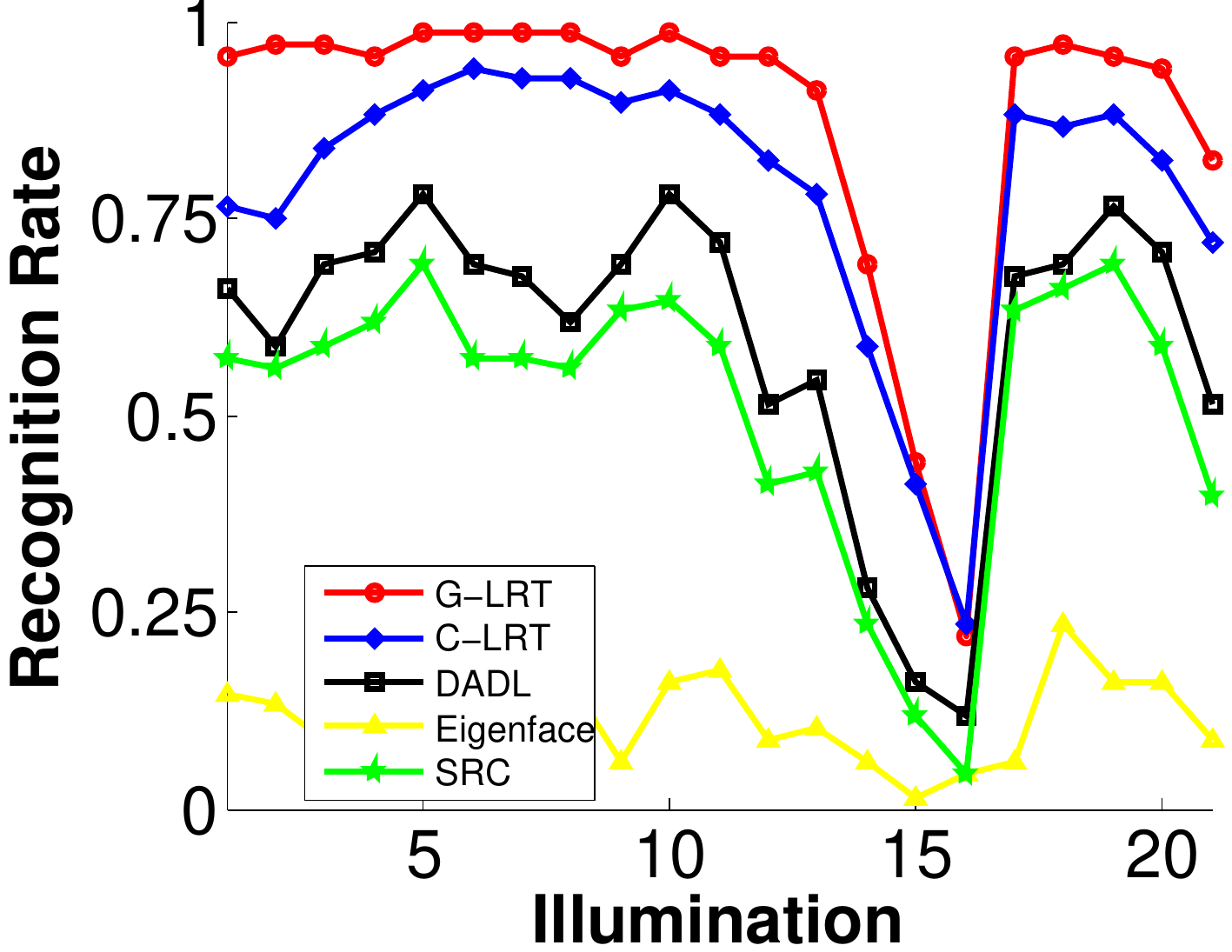}}
\caption{Face recognition accuracy under combined pose and illumination variations on the CMU PIE dataset. The proposed methods are denoted as \emph{G-LRT} in color red and \emph{C-LRT} in color blue. The proposed methods significantly outperform the comparing methods, especially for extreme poses c02 and c14.}
\label{fig:pieacc-eccv}
\end{figure*}

\begin{figure*} [t]
\centering
\subfloat[Globally transformed testing samples for \emph{subject1} ] {\label{fig:PIEShareTest01eccv} \includegraphics[angle=0, height=0.14\textwidth, width=.5\textwidth]{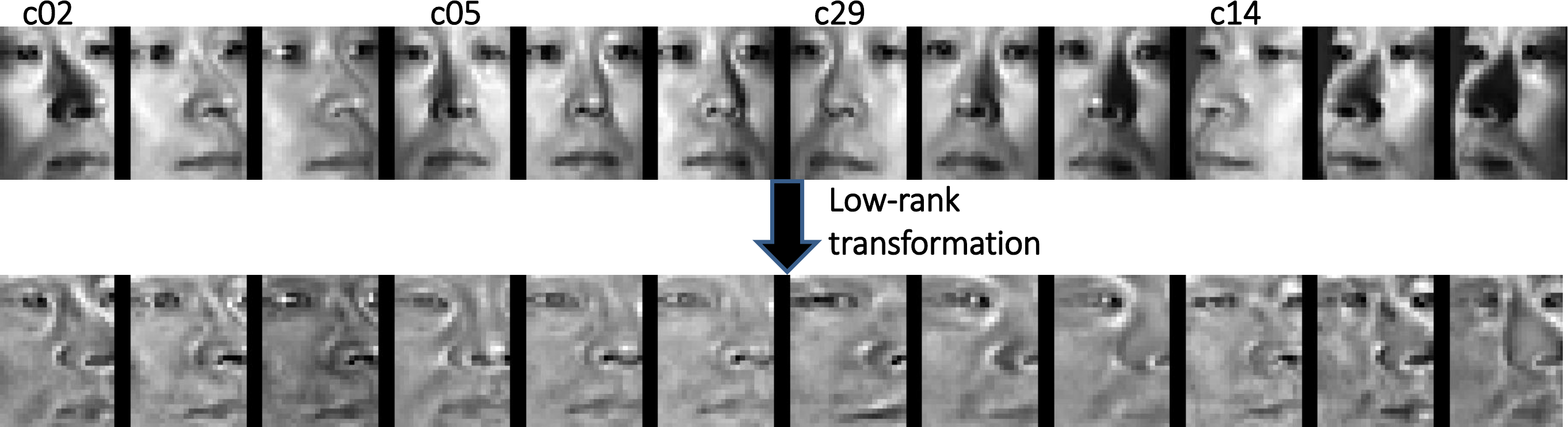} \hspace{0pt}}
\subfloat[Globally transformed testing samples for \emph{subject2} ] {\label{fig:PIEShareTest02eccv} \includegraphics[angle=0, height=0.14\textwidth, width=.5\textwidth]{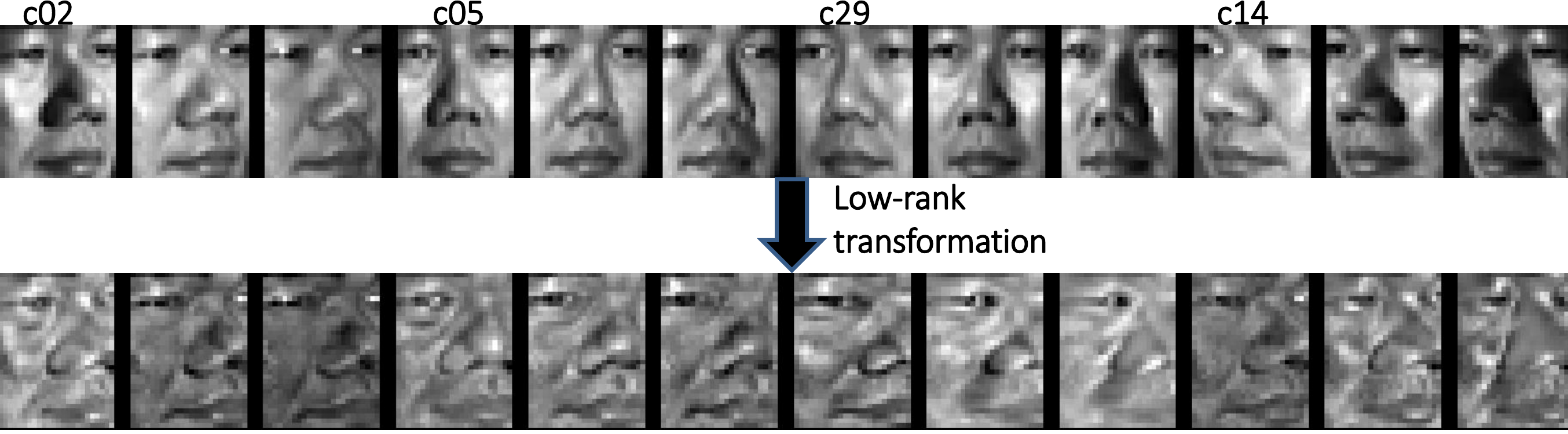}}
\caption{Face recognition under combined pose and illumination variations using global low-rank transformation.}
\label{fig:PIEShareTesteccv}
\end{figure*}

To enable the comparison with \cite{dadl}, we adopt the similar setup in \cite{dadl} for
face recognition under combined pose and illumination variations for the CMU PIE dataset.
We use 68 subjects in 5 poses, c22, c37, c27, c11 and c34,  under 21 illumination conditions for training; and classify 68 subjects in 4 poses, c02, c05, c29 and c14, under 21 illumination conditions.

Three face recognition methods are adopted for comparisons: Eigenfaces \cite{eigenface},
SRC \cite{Wright09}, and DADL \cite{dadl}. SRC is a state of the art method to use sparse representation for face recognition. DADL is an enhanced version of SRC, adapting dictionaries to the actual visual domains.
As shown in Fig.~\ref{fig:pieacc-eccv}, the proposed methods, both the global LRT (G-LRT) and class-based LRT (C-LRT),  significantly outperform the comparing methods, especially for extreme poses c02 and c14.
Some testing examples using a global transformation are shown in Fig.~\ref{fig:PIEShareTesteccv}. We notice that the transformed faces for each subject exhibit reduced variations caused by pose and illumination.

\section{Conclusion}
\label{sec:con}

We presented an approach to learn image low-rank transformations for face recognition under pose and illumination variations. The learned transformations restore for faces from the same subject a low-dimensional structure, which is often violated
 by the change of visual domains, and, at the same time, force a high-rank structure for faces from different subjects for discrimination. Thus, we reduce the variation within the classes and increase separations between the classes to achieve better face recognition across domains.
 The proposed method can be generalized for object recognition, however, further experimental validations are to be performed. We also plan to study in details the mentioned connections with coded aperture design.

{\small
\bibliographystyle{ieee}
\bibliography{reclr}

\begin{thebibliography}{10}\itemsep=-1pt

\bibitem{9point}
R.~Basri and D.~W. Jacobs.
\newblock Lambertian reflectance and linear subspaces.
\newblock {\em IEEE Trans. on Patt. Anal. and Mach. Intell.}, 25(2):218--233,
  February 2003.

\bibitem{rpca}
E.~J. Cand\`{e}s, X.~Li, Y.~Ma, and J.~Wright.
\newblock Robust principal component analysis?
\newblock {\em J. ACM}, 58(3):11:1--11:37, June 2011.

\bibitem{CS1}
W.~R. Carson, M.~Chen, M.~R.~D. Rodrigues, R.~Calderbank, and L.~Carin.
\newblock Communications-inspired projection design with application to
  compressive sensing.
\newblock {\em SIAM J. Imaging Sci.}, 5(4):1185--1212, 2012.

\bibitem{CS2}
J.~D. Carvajalino, G.~Yu, L.~Carin, and G.~Sapiro.
\newblock Task-driven adaptive statistical compressive sensing of gaussian
  mixture models.
\newblock {\em IEEE Trans. Signal Processing}, 2013.

\bibitem{smd}
C.~Castillo and D.~Jacobs.
\newblock Using stereo matching for 2-d face recognition across pose.
\newblock {\em IEEE Trans. on Patt. Analysis and Mach. Intell.}, 31:2298--2304,
  2009.

\bibitem{s-smd}
C.~Castillo and D.~Jacobs.
\newblock Wide-baseline stereo for face recognition with large pose variation.
\newblock In {\em Proc. IEEE Computer Society Conf. on Computer Vision and
  Patt. Recn., Colorado Springs, CO}, June 2011.

\bibitem{DPM}
P.~Felzenszwalb, R.~Girshick, D.~McAllester, and D.~Ramanan.
\newblock Object detection with discriminatively trained part-based models.
\newblock {\em IEEE Trans. on Patt. Anal. and Mach. Intell.}, 32(9):1627--1645,
  2010.

\bibitem{yaleb}
A.~S. Georghiades, P.~N. Belhumeur, and D.~J. Kriegman.
\newblock From few to many: Illumination cone models for face recognition under
  variable lighting and pose.
\newblock {\em IEEE Trans. on Patt. Anal. and Mach. Intell.}, 23(6):643--660,
  June 2001.

\bibitem{lightfield}
R.~Gross, S.~Baker, I.~Matthews, and T.~Kanade.
\newblock Face recognition across pose and illumination.
\newblock In {\em Handbook of Face Recognition}. Springer-Verlag, 2004.

\bibitem{congeal}
G.~B. Huang, V.~Jain, and E.~Learned-Miller.
\newblock Unsupervised joint alignment of complex images.
\newblock In {\em Proc. Intl. Conf. on Computer Vision, Rio de Janeiro,
  Brazil}, Oct. 2007.

\bibitem{lcksvd}
Z.~Jiang, Z.~Lin, and L.~S. Davis.
\newblock Learning a discriminative dictionary for sparse coding via label
  consistent {K-SVD}.
\newblock In {\em Proc. IEEE Computer Society Conf. on Computer Vision and
  Patt. Recn., Colorado springs, CO}, June 2011.

\bibitem{omp}
Y.~C. Pati, R.~Rezaiifar, and P.~S. Krishnaprasad.
\newblock Orthogonal matching pursuit: recursive function approximation with
  applications to wavelet decomposition.
\newblock {\em Proc. 27th Asilomar Conference on Signals, Systems and
  Computers}, pages 40--44, Nov. 1993.

\bibitem{RASL}
Y.~Peng, A.~Ganesh, J.~Wright, W.~Xu, and Y.~Ma.
\newblock {RASL}: Robust alignment by sparse and low-rank decomposition for
  linearly correlated images.
\newblock In {\em IEEE Computer Society Conf. on Computer Vision and Patt.
  Recn., San Francisco, CA}, 2010.

\bibitem{dadl}
Q.~Qiu, V.~Patel, P.~Turaga, and R.~Chellappa.
\newblock Domain adaptive dictionary learning.
\newblock In {\em Proc. European Conference on Computer Vision, Florence,
  Italy}, Oct. 2012.

\bibitem{lrsalient}
X.~Shen and Y.~Wu.
\newblock A unified approach to salient object detection via low rank matrix
  recovery.
\newblock In {\em IEEE Computer Society Conf. on Computer Vision and Patt.
  Recn., Providence, Rhode Island}, June 2012.

\bibitem{pie}
T.~Sim, S.~Baker, and M.~Bsat.
\newblock The {CMU} pose, illumination, and expression ({PIE}) database.
\newblock {\em IEEE Trans. on Patt. Anal. and Mach. Intell.}, 25(12):1615
  --1618, Dec. 2003.

\bibitem{eigenface}
M.~Turk and A.~Pentland.
\newblock Face recognition using eigenfaces.
\newblock In {\em Proc. IEEE Computer Society Conf. on Computer Vision and
  Patt. Recn., Maui, Hawaii}, June 1991.

\bibitem{tensorface1}
M.~A.~O. Vasilescu and D.~Terzopoulos.
\newblock Multilinear analysis of image ensembles: Tensorfaces.
\newblock In {\em Proc. European Conf. on Computer Vision, Copenhagen,
  Denmark}, May 2002.

\bibitem{Viola-Jones}
P.~Viola and M.~Jones.
\newblock Robust real-time face detection.
\newblock {\em International Journal of Computer Vision}, 57:137--154, 2004.

\bibitem{src-align}
A.~Wagner, J.~Wright, A.~Ganesh, Z.~Zhou, H.~Mobahi, and Y.~Ma.
\newblock Toward a practical face recognition system: Robust alignment and
  illumination by sparse representation.
\newblock {\em IEEE Trans. on Patt. Anal. and Mach. Intell.}, 34(2):372--386,
  2012.

\bibitem{subdifferential}
G.~A. Watson.
\newblock Characterization of the subdifferential of some matrix norms.
\newblock {\em Linear Algebra and Applications}, 170:1039--1053, 1992.

\bibitem{Wright09}
J.~Wright, M.~Yang, A.~Ganesh, S.~Sastry, and Y.~Ma.
\newblock Robust face recognition via sparse representation.
\newblock {\em IEEE Trans. on Patt. Anal. and Mach. Intell.}, 31(2):210--227,
  2009.

\bibitem{Zhang10}
Q.~Zhang and B.~Li.
\newblock Discriminative {k-SVD} for dictionary learning in face recognition.
\newblock In {\em Proc. IEEE Computer Society Conf. on Computer Vision and
  Patt. Recn., San Francisco, CA}, June 2010.

\bibitem{TILT}
Z.~Zhang, X.~Liang, A.~Ganesh, and Y.~Ma.
\newblock {TILT}: transform invariant low-rank textures.
\newblock In {\em Proceedings of the 10th Asian conference on Computer vision},
  2011.

\bibitem{posemodel}
X.~Zhu and D.~Ramanan.
\newblock Face detection, pose estimation and landmark localization in the
  wild.
\newblock In {\em IEEE Computer Society Conf. on Computer Vision and Patt.
  Recn., Providence, Rhode Island}, June 2012.

\end{thebibliography}
}

\end{document}